# Uncovering the Origins of Instability in Dynamical Systems: How Attention Mechanism Can Help?

Nooshin Bahador[1,2,4] and Milad Lankarany[1,2,3,4]*

[1] Krembil Research Institute – University Health Network (UHN), Toronto, ON, Canada
[2] Institute of Biomaterials & Biomedical Engineering (IBBME), University of Toronto, Toronto, ON, Canada
[3] Department of Physiology, University of Toronto, Toronto, ON, Canada
[4] KITE Research Institute, Toronto Rehabilitation Institute - University Health Network (UHN), Toronto, ON, Canada

* To whom correspondence should be addressed: Milad Lankarany
The Krembil Research Institute – University Health Network (UHN), 60 Leonard Ave, Toronto, M5T 0S8, Canada

**Abstract**

The behavior of the network and its stability are governed by both dynamics of individual nodes as well as their topological interconnections. Attention mechanism as an integral part of neural network models was initially designed for natural language processing (NLP), and so far, has shown excellent performance in combining dynamics of individual nodes and the coupling strengths between them within a network. Despite undoubted impact of attention mechanism, it is not yet clear why some nodes of a network get higher attention weights. To come up with more explainable solutions, we tried to look at the problem from stability perspective. Based on stability theory, negative connections in a network can create feedback loops or other complex structures by allowing information to flow in the opposite direction. These structures play a critical role in the dynamics of a complex system and can contribute to abnormal synchronization, amplification, or suppression. We hypothesized that those nodes that are involved in organizing such structures can push the entire network into instability modes and therefore need higher attention during analysis. To test this hypothesis, attention mechanism along with spectral and topological stability analyses was performed on a real-world numerical problem, i.e., a linear Multi Input Multi Output state-space model of a piezoelectric tube actuator. The findings of our study suggest that the attention should be directed toward the collective behaviour of imbalanced structures and polarity-driven structural instabilities within the network. The results demonstrated that the nodes receiving more attention cause more instability in the system. Our study provides a proof of concept to understand why perturbing some nodes of a network may cause dramatic changes in the network dynamics.

# 1. Introduction

In many networks, spesific nodes at critical positions within the network act as drivers that push the system into particular modes of action [1]. Observing large-scale network catastrophe in sociological and biological systems like the widespread effects of epilepsy in brain network, poses a few questions of—How does a chaotic regime start in complex networks? Where should be looked for spreading origins, epicentres, or initiators in the network? Which nodes are most influencal in driving changes in network's dynamics? Why these particular nodes have potential ability in facilitating changes in the state of a system? Can imminent shifts be predicted within the network's dynamics prior to onset and enhance preparedness? Answering to these questions motivated us to explore how the local structures within a network cause deteriorating stability and pushing the network into catastrophic regime. This study tried to leverage principles in stability theory and connects them to attention mechanism in neural networks.

Attention mechanism is one of the widely used technique in natural language processing and computer vision which focuses on the most informative parts of the data and significantly improves many processing tasks like image classification, object detection, etc [2]. Attention mechanism can also help the graph convolutional networks to focus on nodes with key contributions in information processing of the graph [3]. In graph neural networks, it has been argued that instead of considering the entire local neighborhood, only nodes with higher attention values should be propagated. According to this assumption, the robustness of the network can be improved by only considering important nodes and ignoring misleading points [4]. Despite the tremendous success of this effective technique, the one thing that still lacks and has not addressed much is an explanation of why attention mechanism works for network analysis and what attention coefficients exactly reflect.

Seeking for an explanation, this study tried to look at this problem from the stability analysis perspective. The stability concept in graph theory looks at how changes in particular node can affect the rest of the network and how the connectivity of that particular node depends on other nodes in the network [5, 6]. Some studies have tried to check the stability properties of graph neural networks to see how changes in the underlying topology can affect the output of network [7]. In terms of model optimization, it has been discussed that unstable nodes in sparse regions of network require to be pulled apart to improve the classification decision [8].

Focusing on stability properties of networks, detection of spreading pathways within the network has been the focus of many recent studies. In cases where abnormalities, chaos or instability can spread rapidly across the network, early spotting the spreading origins is essential to hinder widespread harm. One example of such condition is when a small perturbation within the brain network of epileptic patient leads to seizure propagation at a life-threatening level [9].

A large body of literature has tried to rank the spreading ability of nodes in the network. It has been assumed that the nodes with either high nodal centrality or high betweenness centrality are influential in large-scale spreading [10, 11]. However, this assumption turned to not work for all the real-world networks and there were cases in which the highly connected nodes or the nodes with highest betweenness had little effect on spreading process [12, 13]. One study has argued that the topology of the network organization plays a key role in wide-spread phenomena. This study has also claimed that the spreading process may not necessarily originate in just a single node, but they can start from many nodes simultaneously [14]. There are some reported cases including localized attack on networks where spreading can happen locally by only covering a specific group of nodes [15-19]. Considering all these different reported evidences, this question of how the spreading ability of nodes in the network should be ranked still remained under investigation.

The fact that topological properties of a network affect the dynamical process [20] can suggest that spreading dynamics are rooted in some hidden structures in the network. It has been reported that complex temporal dynamics in real-world networks may be induced by the spatial dimension [21]. Looking at spatial aspects of chaotic dynamics, one study has argued that the dynamics of the system becomes chaotic because of homogeneity breaking [22]. There is also strong evidence that symmetry breaking can cause instabilities in networks [23]. Considering these claims and our initial assumption, we further assumed that existence of hidden symmetry-breaking structures within the network may also cause emergence of spreading dynamics.

Considering the fact that both attention mechanism and stability analysis focus on influential nodes, the final question here is whether unstable nodes are the nodes that need more attention. We address this question in the following sections. The first subsection of Method section describes the case study which is a real-world numerical problem. This problem is mapped from state-space model into graph representation for further analysis. The states of model are considered as individual nodes within the network. In subsection 2, an attention-enhanced graph convolutional network (AGCN) is used to classify the nodes of this network. After learning process of the AGCN, an attention coefficient for each pair of nodes is extracted and nodes with higher attention coefficients are identified. To check our hypothesis, which stated that those nodes that have potential to move the entire network into the unstable mode are the nodes that have higher attention coefficients, three different stability analyses are performed. Finally, the nodes with higher instability risk are identified and compared to those with higher attention coefficients.

## 2. Method

### 2.1. Simulated Dynamical System

Dynamical systems can be stabilized by state feedback which involves using the state vector for controling system dynamics. This feedback mechanism can be applied to controllable states. Identifying most important states can be very helpful in designing an optimum closed-loop control system. This study tried to identify important states using attention mechanism and proves that these important nodes are the ones that show more tendency toward instability.

One of the dynamical systems that requires feedback mechanism to reach stability is Piezoelectric tube actuators. The problem of modeling of these actuators has been considered as a real-world numerical example in this study. These actuators are frequently used in micro/nano-scales applications and they are highly sensitive to the uncertainties including environmental variations. The piezoelectric tube actuator can be expressed by a linear Multi Input Multi Output state-space model using following equations [24]:

$$\begin{cases} \dot{x}(t) = Ax(t) + Bu(t) \\ y(t) = Cx(t) + Du(t) \end{cases} \quad (1)$$

where, A, B, C and D are respectively state matrix, input matrix, output matrix and feedforward matrix. Variables $x$ and $y$ are respectively state and output vectors.

$$A = \begin{bmatrix} 0 & 1 & 0 & 0 & 0 & 1 & 0 & 0 \\ 0 & 0 & 1 & 0 & 0 & 0 & 1 & 0 \\ 0 & 0 & 0 & 1 & 0 & 0 & 0 & 1 \\ -a_{14} & -a_{13} & -a_{12} & -a_{11} & 0 & 0 & -a_{22} & -a_{21} \\ 0 & 1 & 0 & 0 & 0 & 1 & 0 & 0 \\ 0 & 0 & 1 & 0 & 0 & 0 & 1 & 0 \\ 0 & 0 & 0 & 1 & 0 & 0 & 0 & 1 \\ 0 & 0 & -a_{32} & -a_{31} & -a_{44} & -a_{43} & -a_{42} & -a_{41} \end{bmatrix}$$

The values of coefficients in matrix A are listed in Table 1.

Table 1. The coefficients of matrix A

| | |
|---|---|
| $a_{11} = 517.0544$ | $a_{31} = 191.8224$ |
| $a_{12} = 4.2614$ | $a_{32} = 49.4899$ |
| $a_{13} = 1.3083$ | $a_{41} = 239.0092$ |
| $a_{14} = 2.7480$ | $a_{42} = 1.7819$ |
| $a_{21} = 0.9492$ | $a_{43} = 3.6549$ |
| $a_{22} = 2.6331$ | $a_{44} = 5.2346$ |

Figure 1 shows a graph that was created from the dynamical system (1), considering A as an adjacency matrix representing the topology of a network.

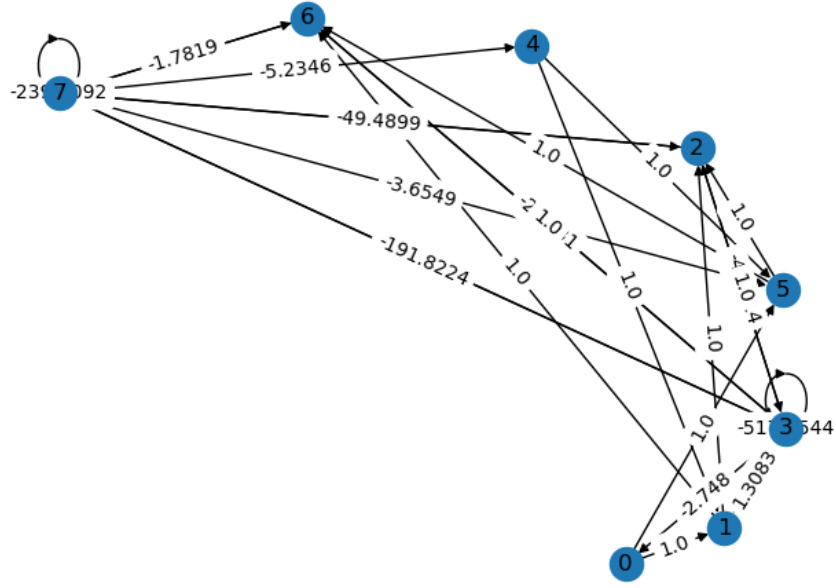

**Figure 1**. Graph representation of dynamical system in (1) by considering A as the adjacency matrix. Matrix A represents dynamics of hidden states in piezoelectric tube actuator model and each node corresponds to one state. In this study, the output vector y, as well as variables C and D, has no impact on the graph shown in this figure.

A random feature set consisting of three samples was assigned to each node according to Table 2.

Table 2. The random feature set assigned to each node as attributes

| Node Id | Feature Set | | |
|---|---|---|---|
| 0 | 0.5 | -0.1 | 0.3 |
| 1 | 0.2 | 0.1 | 0.7 |
| 2 | -0.5 | 0.7 | -0.1 |
| 3 | -0.1 | -0.6 | 0.4 |
| 4 | 0.3 | -0.5 | -0.2 |
| 5 | 0.1 | -0.1 | -0.4 |
| 6 | 0.3 | 0.8 | -0.1 |
| 7 | 0.1 | -0.2 | 0.2 |

## 2.2. Attention Mechanism

In this study, an attention-enhanced graph convolutional network (AGCN), including different modules, was used for node classification. These modules are explained in the following sub-sections.

### 2.2.1. First module: Initial node features embedding

The first module performs a self-attention operator on the nodes, which is a simple dot product (multiplying node features matrix by its transpose) that helps us to represent the relationship among features. The intuition behind self-attention operator is to express how two feature vectors are related in the input space. In this operation, a weighted average over all the input vectors is taken. A visual illustration of this weighted average is shown in Figure 2. The dot product over each pair of feature vectors gives their corresponding weights. If the sign of a feature match with the other one, this weight gets a positive term and if the sign does not match, the corresponding weight is negative. The magnitude of the weight indicates how much the feature should contribute to the total score. As weight value produced by this self-attention operator lies anywhere between negative and positive infinity, both Leaky ReLU and Softmax operators need to be applied to map all the weight values between zero and one and their summation to be one.

$$X = \{\vec{x_1}, \vec{x_2}, ...\}, \vec{x_i} \in \mathbb{R}^N \quad (2)$$
$$\omega_{self} = X \cdot X^T$$
$$\omega_{self} = \{\omega_{11}, \omega_{12}, ...\}$$
$$Y = \omega_{self} X$$
$$Y = \{\vec{y_1}, \vec{y_2}, ...\}, \vec{y_i} \in \mathbb{R}^N$$
$$Y' = Softmax(LeakyRelu(Y))$$
$$Y' = \{\vec{y_1'}, \vec{y_2'}, ...\}$$

where X is the matrix of nodes features. $\omega_{self}$ indicates self-attention weights. Y is the weighted average of node features. $Y'$ is the weighted average of features passed through activation functions.

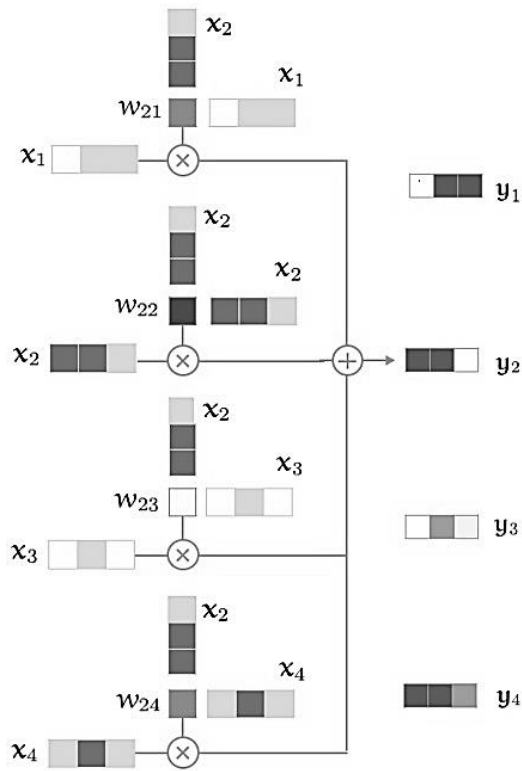

**Figure 2**. Self-attention operator for four sample nodes (figure from [25]). The outputs of $\{y_1, ..., y_i\}$ are aggregates of interactions between inputs of $\{x_1, ..., x_i\}$ and their attention scores of $\{\omega_{11}, \omega_{12} ..., \omega_{ij}\}$.

This weighted average of node features produces a new set of node features as the output of self-attention operator which forms the inputs for the next module.

### 2.2.2. Second module: Learnable attention mechanism

The second module is a single-layer feedforward neural network parameterized by attention weight vector ($\omega_{Att}$). In this module, the feature vectors of each pair in new set of nodes (produced in previous module) are concatenated and passed through Leaky ReLU and SoftMax operators. A visual illustration of this process is shown in Figure 3. The goal here is to extract attention coefficient for each pair of nodes which represents the importance of one node's feature to the feature of another one.

$$\alpha_{ij} = Softmax\left(LeakyRelu\left(\omega_{Att} \cdot \left(\vec{y'_i} \parallel \vec{y'_j}\right)\right)\right) \tag{3}$$

$$\omega_{Att} = \{\omega_{Att_1}, \omega_{Att_2}, \dots\}, \; \vec{\omega_{Att_i}} \in \mathbb{R}^{2N}$$
$$\omega_\alpha = \{\alpha_{12}, \alpha_{13}, \dots\}, \; \vec{\alpha_{ij}} \in (\mathbb{R}^N \times \mathbb{R}^N)$$

where $\omega_{Att}$ is attention weight vector. $y'$ is the weighted average of features passed through activation functions. $\omega_\alpha$ is attention coefficient matrix.

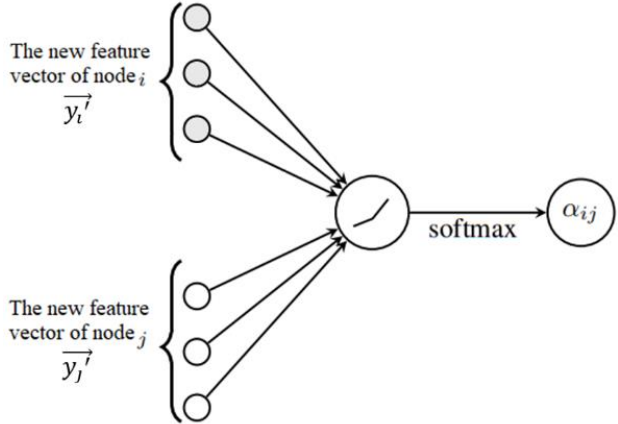

**Figure 3**. Learnable attention mechanism (figure was adjusted from [26]). The weighted average of features corresponds to each pair of nodes are fed into a fully connected layer and passed through a SoftMax activation function to produce attention coefficient.

### 2.2.3. Third module: Graph convolution

The third module perform features aggregation from neighbors of each node. This can be calculated by multiplication of adjacency and features matrices. It should be considered that the features of the node itself is as important as its neighbors. To consider features of the node itself, an identity matrix needs to be added to adjacency matrix (A) to get a new adjacency matrix ($\tilde{A}$). To prevent exploding/vanishing gradients because of high-degree/low-degree nodes and to reduce the sensitivity of network to the scale of input data, the matrix multiplication needs to be scaled according to the node degrees (scaling by both rows and columns). This scaling places more weights on the low-degree nodes and reduce the impact of nodes with high degree. The motivation behind this scaling is that nodes with low degrees have greater influences on their neighbors, whereas nodes with high degrees have lower effects as they spread their influence on too many neighbors. As scaling is done twice (once across rows and once across columns), the square root of the node degree is taken into account. The influence of one node feature on the other nodes can be also reflected by the dot product of new adjacency matrix with attention coefficients matrix. Finally, graph convolution will be completed by put all these modules together and forms a forward model with learnable weight matrix of W.

$$\tilde{A} = A + I \quad (4)$$
$$\tilde{D} = \sum_j \tilde{A}_{ij}$$
$$\hat{A} = \tilde{D}^{-1/2} \tilde{A} \tilde{D}^{-1/2}$$
$$Y" = Softmax\left(LeakyRelu\left((\hat{A} \cdot \omega_\alpha) X W\right)\right)$$

where D is the degree matrix. $\hat{A}$ is the normalized adjacency matrix with added self-loops.

### 2.2.4. Final module: Back propagation and training

The goal here with using back propagation is to update each weight in attention layer (matrix of $\omega_{Att}$) and convolution layer (matrix of W) so that actual output gets closer to the target output. To do this, the partial derivative of error (gradient) with respect to these weights are calculated. It should be considered that the partial derivative of the SoftMax function is the output×(1 – output).

$$W_{(n+1)} = W_{(n)} + \mu\left((y_{target} - y")^T \left((\hat{A} \cdot \omega_\alpha)X(1 - (\hat{A} \cdot \omega_\alpha)X)\right)\right)^T \quad (5)$$
$$z = \|\left(\left((y'(1-y')) \cdot \left(\mu(y_{target} - y")^T\right)\right) X\right), \quad \| \rightarrow Concatenation$$
$$\omega_{Att(n+1)} = \omega_{Att(n)} + \tilde{z}$$

where W is a learnable weight matrix. $\tilde{A}$ is the new adjacency matrix, $\omega_{Att}$ is attention weight vector in single-layer feedforward neural network. $\omega_\alpha$ is attention coefficient matrix for each pair of nodes, $y_{target}$ is target output and $y"$ is the actual output.

### 2.3. Spectral Stability Analysis

Spectral stability of a network is governed by the largest negative eigenvalue of its adjacency matrix [27]. Our hypothesis is that nodes that need more attention are the ones that can push the entire network into the unstable mode. To test our hypothesis and to check the effect of each node on stability of network, we looked at how the perturbation in one column of adjacency matrix [9] reflected in its largest eigenvalue. The perturbation level was initially set to 0.5 and gradually increased to 3. The following matrix shows the resulting adjacency matrix after perturbing node 1 by Δ. Those nodes for which the largest negative eigenvalue of matrix $\hat{A}_2$ moves towards zero, while their perturbation level increases, have potential to push the entire network into the unstable mode.

$$\hat{A}_2 = \begin{bmatrix} 0 & 1+\Delta & 0 & 0 & 0 & 1 & 0 & 0 \\ 0 & 0 & 1 & 0 & 0 & 0 & 1 & 0 \\ 0 & 0 & 0 & 1 & 0 & 0 & 0 & 1 \\ -a_{14} & -a_{13}+\Delta & -a_{12} & -a_{11} & 0 & 0 & -a_{22} & -a_{21} \\ 0 & 1+\Delta & 0 & 0 & 0 & 1 & 0 & 0 \\ 0 & 0 & 1 & 0 & 0 & 0 & 1 & 0 \\ 0 & 0 & 0 & 1 & 0 & 0 & 0 & 1 \\ 0 & 0 & -a_{32} & -a_{31} & -a_{44} & -a_{43} & -a_{42} & -a_{41} \end{bmatrix}$$

where Δ is perturbation level.

### 2.4. Topological Stability Analysis

How the connections with positive and negative signs are arranged within the network? And, how such arrangements affect network stability? Positive and negative signs are respectively referred to the synchronous and anti-synchronous correlation. According to structural balance theory [28], stability of a three-entity system can be investigated by signed association between two entities in the presence of a third party. This could be generalized to any signed network by considering the associations between its motifs/subgraphs and the signed links within the motifs. A motif is a recurring pattern of interconnections within the graph, formed by a subset of nodes with a path between each pair of nodes. Collective behaviour of the imbalanced motifs may push the network towards an unstable state. Considering all possible ways to connect, a motif is structurally imbalanced when the multiplication of signs on its edges turns negative. In a signed graph, counting the number of imbalanced motifs can tell us about the stability of the network. Figure 4 shows some examples of imbalanced arrangement.

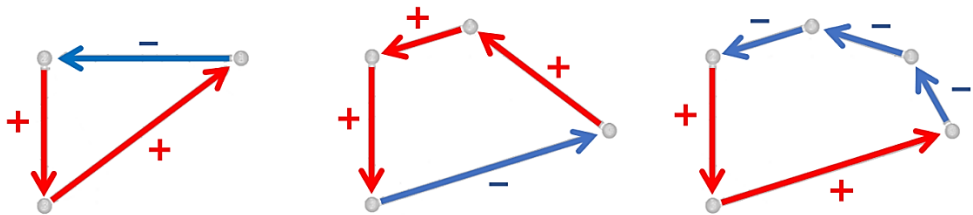

**Figure 4**. Examples of imbalanced motifs with different orders. According to the structural balance theory, loss of balance can occur when the multiplication of the signs of one cycle gets negative.

The influence of each node on the stability of network can be determined by the number of times that a node appears in the imbalanced motifs. To better quantify this influence, a measure is defined that not only considers the imbalanced motifs with different orders, but also it considers the weights of paths which form a cycle within these motifs. For each node and for each imbalanced motif of size 3 that includes that node, the weights of the paths are multiplied and then added together. Same procedure is repeated for the imbalanced motifs of size 4, 5 and 6. The cube root of absolute value for the multiplication of these three calculations are then calculated and the total cost associated with that node is obtained as follow:

$$C_{TN} = \sqrt[3]{\left| W\binom{G_N}{3} \times W\binom{G_N}{4} \times W\binom{G_N}{5} \times W\binom{G_N}{6} \right|} \tag{6}$$

$$W\binom{G_N}{3} = \left( \frac{\sum_{\{i,j,k\}} \omega_{ij}\omega_{jk}\omega_{ki}}{D^2} \right)$$

$$W\binom{G_N}{4} = \left( \frac{\sum_{\{i,j,k,m\}} \omega_{ij}\omega_{jk}\omega_{km}\omega_{mi}}{D^2} \right)$$

$$W\binom{G_N}{5} = \left( \frac{\sum_{\{i,j,k,m,n\}} \omega_{ij}\omega_{jk}\omega_{km}\omega_{mn}\omega_{ni}}{D^2} \right)$$

$$W\binom{G_N}{6} = \left( \frac{\sum_{\{i,j,k,m,n,h\}} \omega_{ij}\omega_{jk}\omega_{km}\omega_{mn}\omega_{nh}\omega_{hi}}{D^2} \right)$$

where $\omega$ is the weight of path between each pair of nodes within an imbalanced motif. The terms of $\binom{G_N}{3}$, $\binom{G_N}{4}$, $\binom{G_N}{5}$ and $\binom{G_N}{6}$ respectively refer to the subset of all possible imbalanced motifs of size 3, 4, 5 and 6 that include one specific node. D is the degree of corresponding node. W corresponds to the normalised sum over the products of motif-paths calculated for each node.

### 2.5. Symmetry-Breaking Stability Analysis

In complex networks, symmetry-breaking means that some nodes attract or transmit flow of information more than other nodes due to the network dynamics or presence of external stimuli. This can lead to the emergence of instability within a network. This phenomenon can occur through the process of self-organization when the nodes in a network interact in a way that they form specific patterns or structures [29]. If a network experiences symmetry-breaking, some nodes may begin to differentiate from other nodes and form distinct sub-networks. This process of differentiation can be thought of as a bifurcation, as it represents a sudden and significant change in the structure and behavior of the network. Occurrence of symmetry-breaking can be seen in nature, for example, when vascular systems like river basins evolves [30]. This process of differentiation can trigger a cascade of further differentiations within those sub-networks. As the differentiations continue to cascade through the network, they can lead to the emergence of chaotic regime.

As network dysfunction can be a function of the microscale structures and flow distributions [31], and spatial symmetry-breaking is one way of studying patterns of information flow, this subsection aimed at identifying spreaders of instability in the network by exploring spatial symmetry-breaking behavior in local flow structures.

Inspired by Flabellate antenna (Figure 5), more than two paths can be branched off from each bifurcation point. They are called Flabellate-shaped bifurcation in this study. Depending on the polarity and strength of individual connections within this symmetry-braking structure, a polarity transition can occur to form a fractal dipole (Figure 6). This topological polarity transition breaks the balance and has the potential of spreading instability across the network.

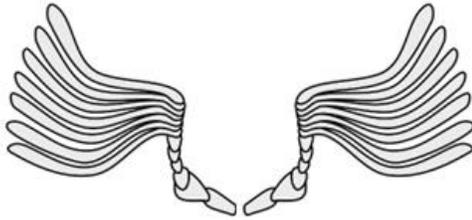
**Figure 5**. Flabellate antenna (adjusted from [32]), shaped in the form of a tree trunk giving rise to multiple branches.

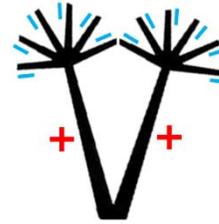
**Figure 6**. Fractal dipole. In each bifurcation point, multiple paths are branched off and polarity is reversed.

To find bifurcation nodes in a network where symmetry breaking along with polarity transition occurs, first, hidden structure of information flow needs to be extracted. As topological properties of a system affect its dynamics, extracting hidden information flow structures in the network provides a useful tool to understand the dynamical behaviour of the network. Graph-based random walk is one of the well-known algorithms, inspired by natural language processing, that can reveal these local structures of information flow [33]. Walking on the graph means moving from one node to another in the direction of the edge, and the flow of information within the network corresponds to walker stepping between nodes. In addition to information flow, activity dynamics on networks can be also modeled by graph-based random walk [34]. Considering that the random walk on network can model information spreading and capture network dynamics [35], we leveraged from graph-based random walk algorithm to investigate the existence of symmetry-breaking structures that are not visible in the network and ranked the nodes of network based on their ability in pushing the network into unstable modes. These random walks represent local structure of information flow distribution and show how information from one node spreads to the other neighboring nodes.

Our goal is to understand whether hidden local structures of information flow can push the network into unstable modes? We hypothesized that an emergence of local polarized flabellate-shaped bifurcation in information flow pathway causes symmetry breaking and identifies initiator of instability within the network.

Each division of bifurcation can branch off in the form of nested projections accompanied by polarity transition. These polarized structures of information flow with fractal-like geometry tend to propagate perturbation faster across the network.

Inspired by the formula for electric dipole moment for a pair of charges which is computed based on the magnitude of charges multiplied by the distance between them, a measure was introduced to represent the overall moment generated by the potential symmetric fractal dipole. The individual nodes within the graph are considered as charges with unit magnitude, and the edge weight represents the distance between two charges. In this study, this measure was called the normalized summation of transition cost (NSTC). Given an array of weights of traversed edges in each two-step random walk starting from node k, the product of edge weights corresponding to each path is computed. All the products of path's weights traversed from each node are then summed up together and normalized by dividing by $N_k$, where k is the index of starting node and N is total number of paths traversed from the starting node. The normalized summation of transition cost as a measure of the overall moment generated by the potential symmetric fractal dipole:

$$NSTC_k = \frac{\Sigma \Pi_{N_k} \omega_k}{N_k} \quad , \quad \omega_k = \{\omega_{ki}, \omega_{ij}\} \tag{7}$$
$$NSTC_k = \frac{\sum_{k=0}^{n}(\omega_{ki} \times \omega_{ij})}{N_k}$$

where $k$ is starting node, $i$ is the visited node in the first step, and $j$ is the visited node in the second step.

The more the $NSTC_k$ is negative the stronger the topological polarity transition is. Nodes become more unstable given stronger topological polarity transition . Unstable nodes have higher potential to spread the instability across the network. The spreading ability of nodes are ranked based on the negativity of $NSTC_k$.

## 3. Results

Our hypothesis was tested on the state-space model of actuator, represented by equation (1). First, attention coefficients were extracted for all the nodes using AGCN. Then, three different stability analyses were performed and the nodes with higher instability risk were identified in each analysis. These three stability analyses included: 1- spectral stability analysis, 2- topological stability analysis, and 3- symmetry-breaking stability analysis.

### 3.1. Attention Mechanism

Considering the connections between nodes in the Figure 1, nodes of 0, 3, 4 and 7 form one cluster (4-degree nodes), and nodes of 1, 2, 5 and 6 form another cluster (2-degree nodes). Two different scenarios were tested. In first scenario, a AGCN model was trained to classify these two clusters. In second scenario, perturbation on feature set of node 0 was applied and a AGCN model was trained to classify these two clusters in the presence of node feature perturbation. The perturbation of feature set was done by multiplying a factor of 2. The labels of 0.01 and 0.2 for the first and second clusters were respectively assigned.

Figures 7 and 8 show the training loss as a function of iteration numbers for two scenarios, namely, without and with perturbation. In both scenarios, the training loss approximately converged to loss value of 0.0028 after 500 iterations, confirming the robustness of AGCN model to feature perturbation.

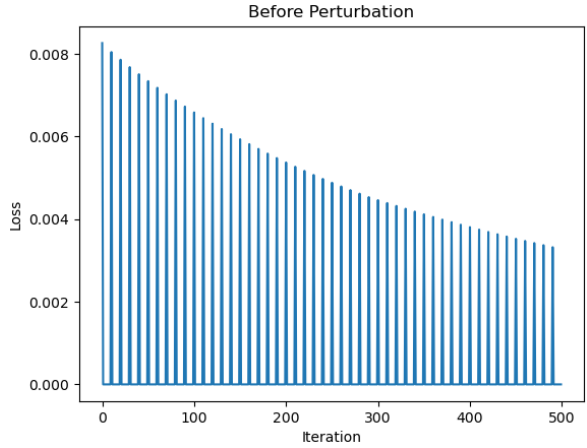
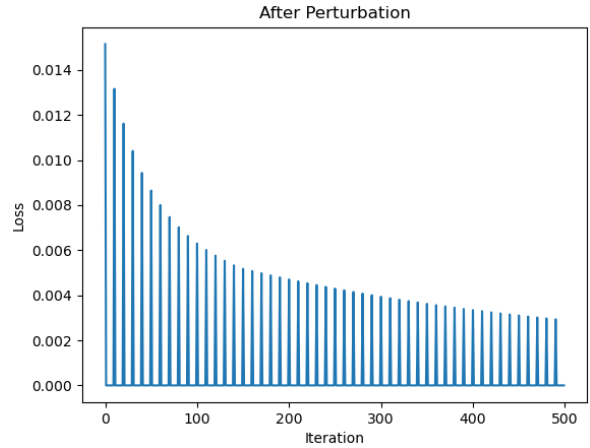

**Figure 7**. Training loss for the scenario without perturbation

**Figure 8**. Training loss for the scenario with perturbation

Table 3 compares the model predictions against truth labels for above-mentioned scenarios. The model predictions summarized in Table 3 were not affected by feature perturbation of node 1, indicating the robustness of AGCN model with respect to feature perturbation. Figure 9 shows nodes of 2 and 6 have the highest attention coefficients.

Table 3. Model prediction of node labels

| Node Id | Actual Label | Predicted Output | |
|---|---|---|---|
| | | without perturbation | with perturbation |
| 0 | 0.01 | 0.083 | 0.0789 |
| 1 | 0.20 | 0.186 | 0.183 |
| 2 | 0.20 | 0.146 | 0.162 |
| 3 | 0.01 | 0.081 | 0.078 |
| 4 | 0.01 | 0.082 | 0.080 |
| 5 | 0.20 | 0.157 | 0.165 |
| 6 | 0.20 | 0.178 | 0.170 |
| 7 | 0.01 | 0.082 | 0.080 |

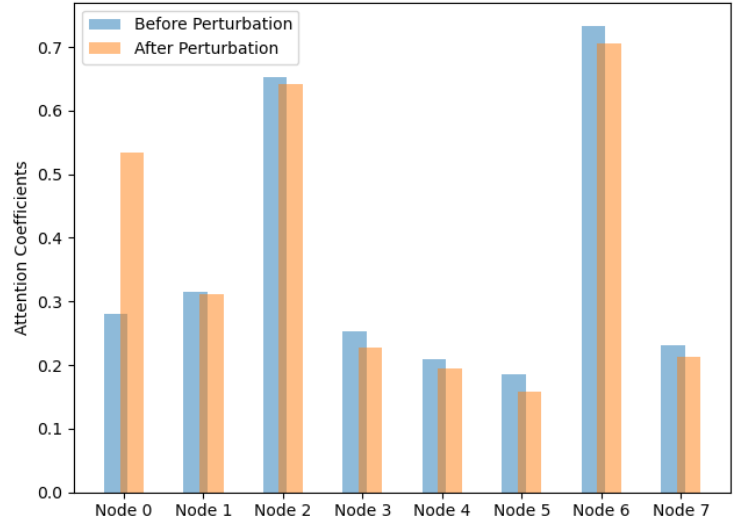

**Figure 9**. Comparing attention coefficients of each node for the scenario with and without perturbation on Node 0. Nodes of 2 and 6 are the ones that need more attention.

### 3.2. Spectral Stability Analysis

To test our hypothesis and check the effect of each node on stability of network, we looked at how the perturbation in one column of adjacency matrix [9] reflected in its largest eigenvalue. To verify the need of unstable nodes for more attention, we performed spectral stability analysis and calculated the change in largest eigenvalue of adjacency matrix by increasing the perturbation level. Figure 10 shows how different nodes in the graph responded to increase in perturbation level. As seen in Figure 10, nodes of 2 and 6 are those that may move the system towards instability because the largest eigenvalue get closer to zero as perturbation level on these nodes increases.

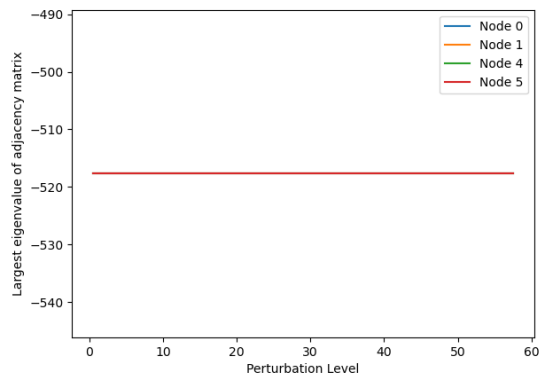

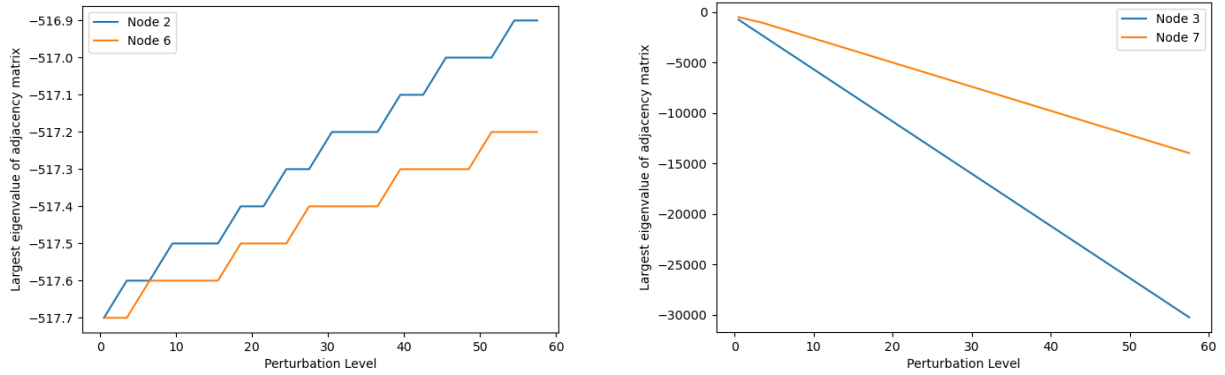

**Figure 10**. The effect of structural perturbation on the largest eigenvalue of adjacency matrix. The perturbation level was initially set to 0.5 and gradually increased by step of 3. Driving to zero by increasing the perturbation level only occurred for the largest eigenvalues of nodes of 2 and 6.

### 3.3. Topological Stability Analysis

To check to what extent unstable nodes involve in imbalanced motifs within the networks, topological stability analysis was performed. The goal was to detect those nodes that lie within the path of imbalanced motifs of different orders. Figure 11 shows the trajectory starts at node 2 and traverses within three sample motifs of different order.

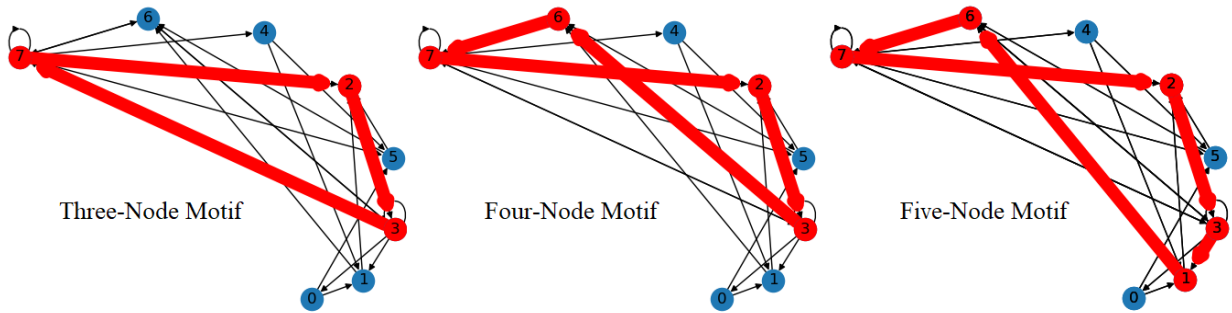

**Figure 11**. N-node motif within the network under study

In the network under study, all the imbalanced motifs of size 3 which passed a specific node were first extracted. The product of weights of the paths within each motif were then performed and stored as a single score. Similar scores were computed for other motifs that passed the same node, and all these scores were summed up to obtain the total score for each node. The total score of each node was normalised based on square of node's degree. Similar procedure was repeated for imbalanced motifs of size 4, 5 and 6. Table 4 summarized the total scores for each individual nodes and for each order of motif. The last column of Table 4 shows the total cost obtained from multiplication of these three scores and taking the cube root of absolute value of it. Figure 12 provides a visual representation of total cost for each node and reflects the potential role of node 2 and 6 in moving network into unstable mode.

Table. 4. Score associated with imbalanced motif-paths traversed from each node

| Node | Three-Node Motif | Four-Node Motif | Five-Node Motif | Six-Node Motif | Total Cost |
|------|------------------|-----------------|-----------------|----------------|------------|
| 0    | 0.00             | -1.22           | 0.00            | 0.00           | 0.00       |
| 1    | 0.00             | -20.71          | -2.68           | -0.38          | 0.00       |
| 2    | -0.10            | -7.41           | -1.86           | -0.26          | 0.71       |
| 3    | 0.00             | -5.12           | -0.67           | -0.09          | 0.00       |
| 4    | 0.00             | -2.32           | 0.00            | 0.00           | 0.00       |
| 5    | -0.29            | -0.63           | 0.00            | -0.38          | 0.00       |
| 6    | -0.10            | -7.41           | -1.86           | -0.26          | 0.71       |
| 7    | -0.07            | -5.22           | -0.67           | -0.09          | 0.29       |

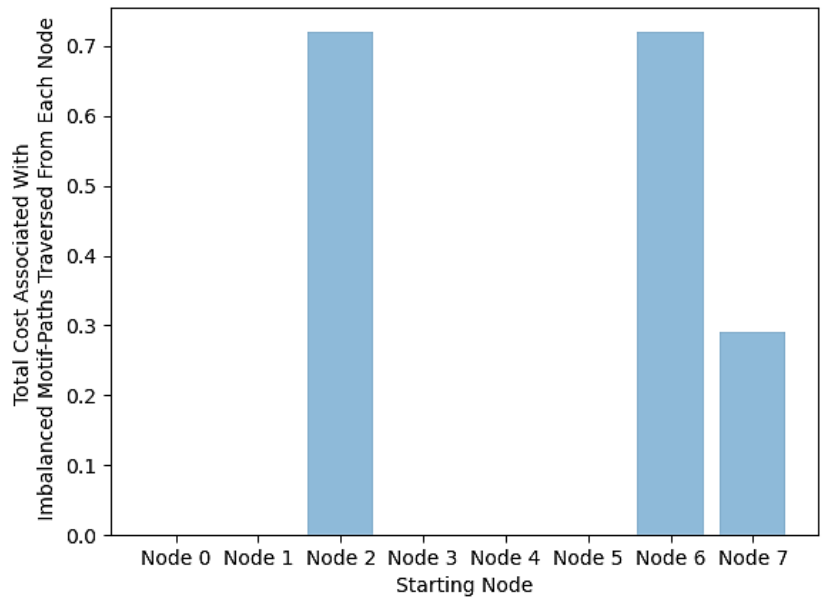

**Figure 12**. The contribution of each node in froming imbalanced motifs with different degrees. This contribution shows the overal influence of each node on moving network toward an unstable state.

### 3.4. Symmetry-Breaking Stability Analysis

To confirm whether the unstable nodes contribute to some polarized structures within the network, the symmetry-breaking stability analysis was performed. To do this, the local structure of information flow distribution was extracted for each node. The process of extracting these information flow distributions for two single nodes has been plotted in Figure 13 and 14. Figure 13 shows all the paths start at node 0 and traverses within two-step random walk. Similar figure has plotted for random walk starting from node 2 (Figure 14).

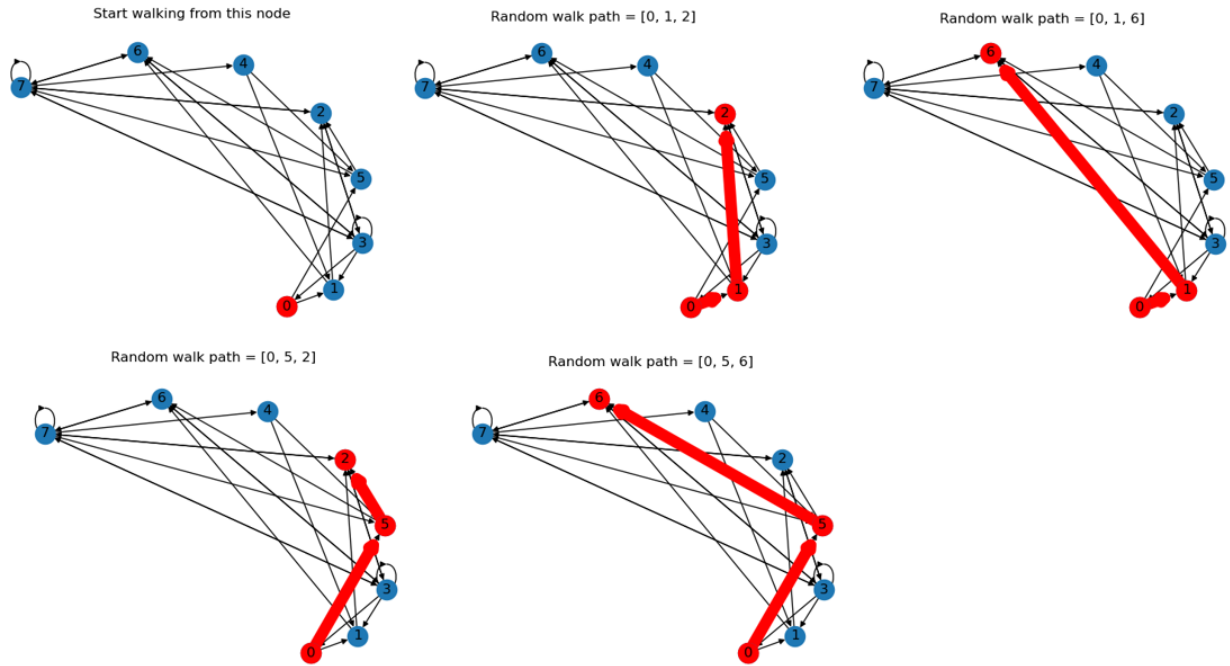

**Figure 13**. Visualisation of random walk starting from node 0. These two-step random walks are all the possible paths initiated from node 0.

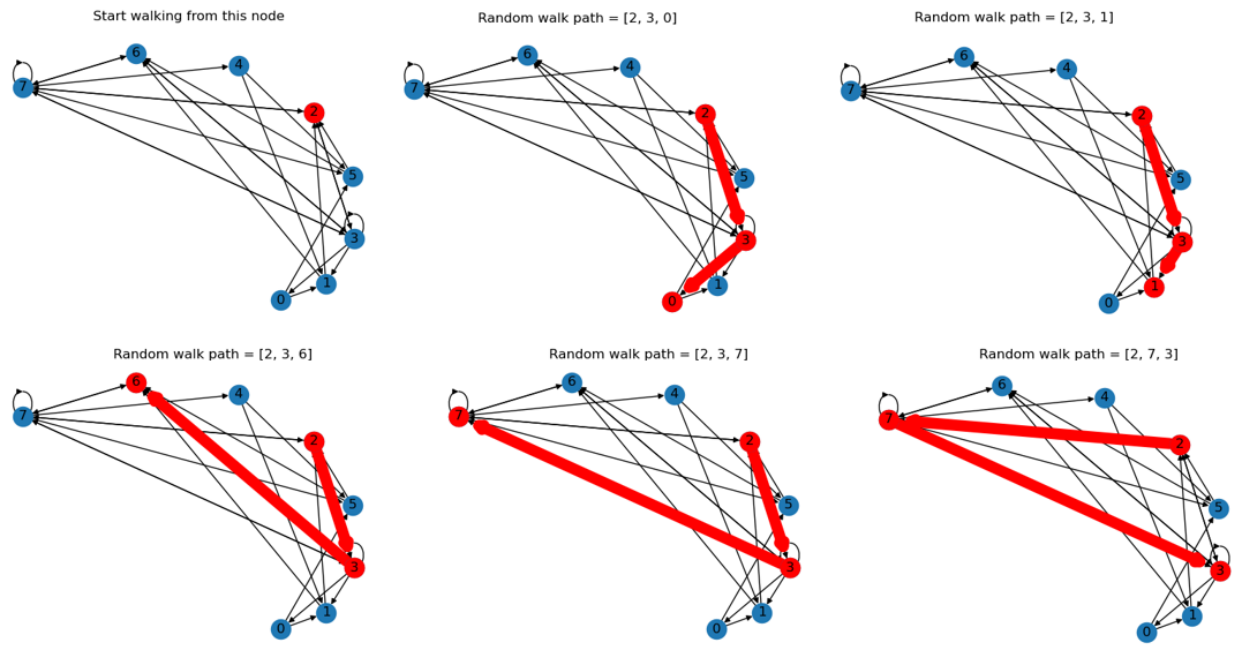

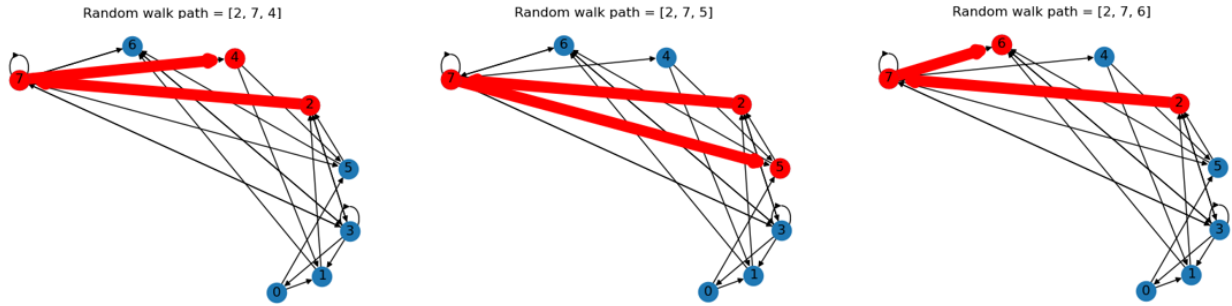

**Figure 14**. Visualisation of random walk starting from node 2.

By simontanously plotting all the random walks coresponding to each node (Figure 15), a clear pattern of flabellate-shaped bifurcation appeared on nodes 2 and 6.

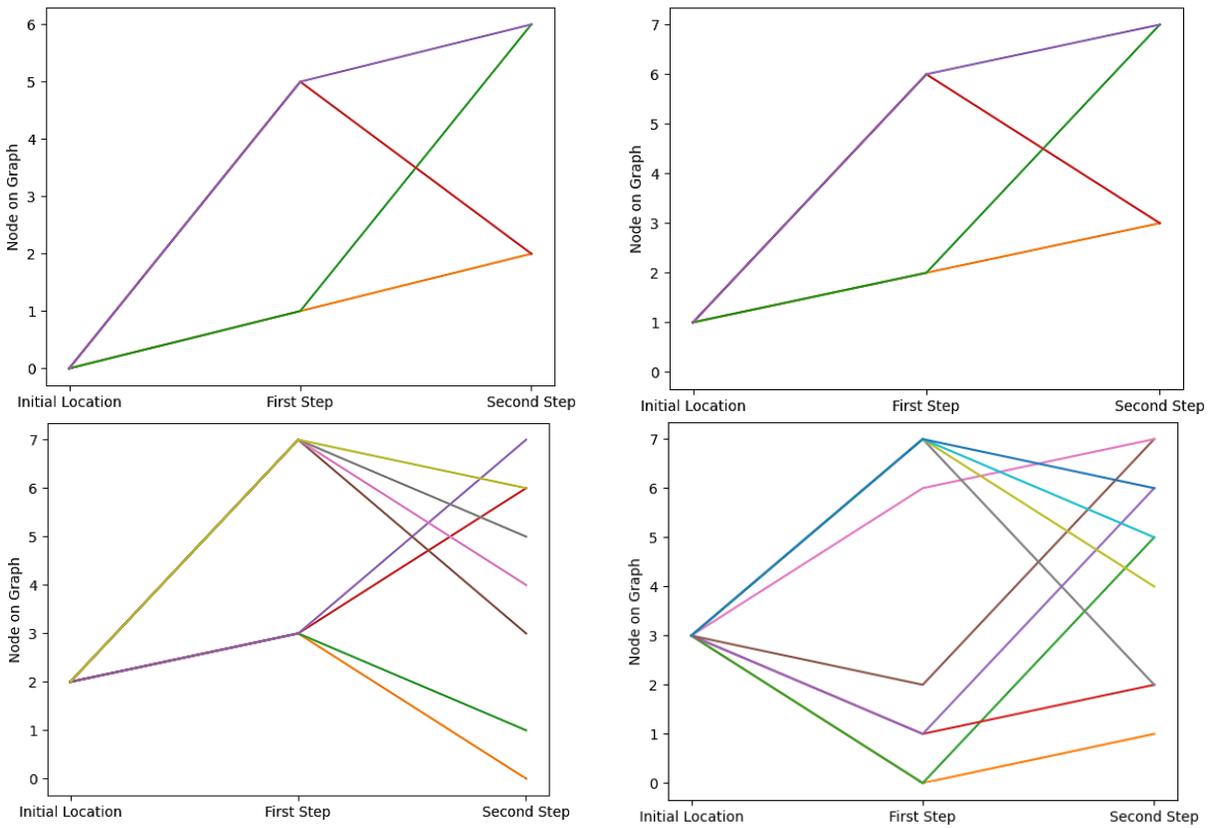

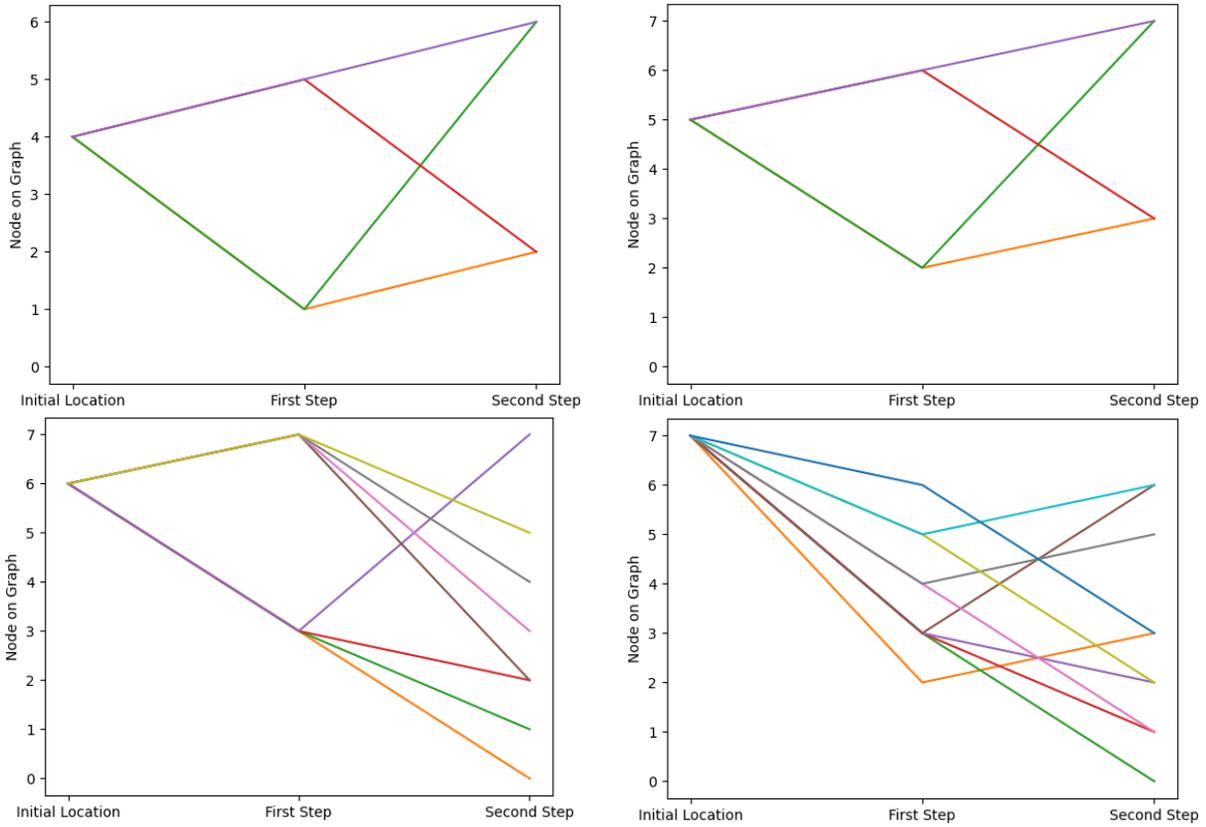

**Figure 15**. Information flow tree rooted in each node. Flabellate-shaped bifurcations observed in nodes of 2 and 6. To determine whether these bifurcations form a fractal dipole, polarity transition should be checked.

As shown in Figure 16, nodes 2 and 6 are those influential spreaders being able to push the network into unstable mode.

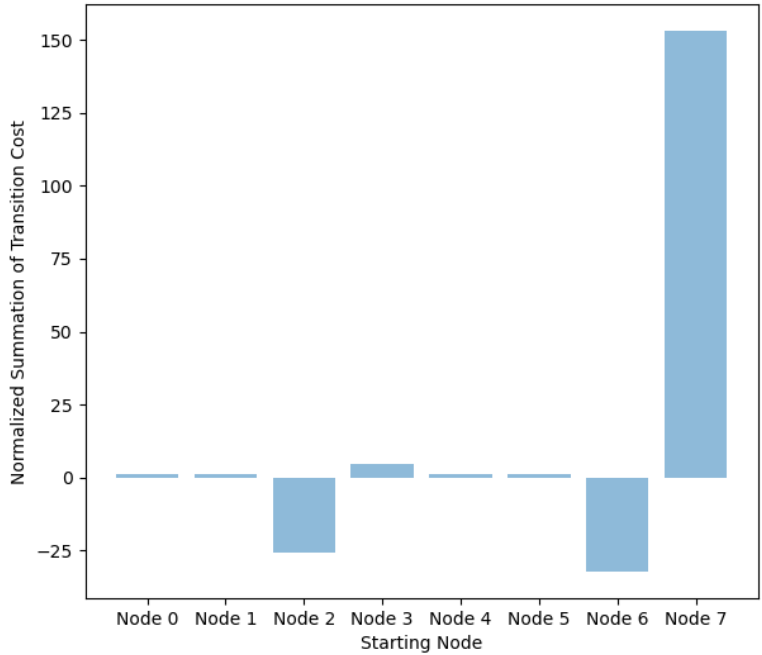

**Figure 16**. *NSTC* as the measure of spreading ability of each node. The negative score of NSTC corresponds to the node where topological polarity transition occures. Those nodes with more negative values of NSTC have higher ability to spread the instability across the network.

## 4. Discussion and Conclusion

This study provided a proof of concept for triangular relationships between attention mechanism, instability, and structural dynamics in the network. We showed that the mechanism that enables a machine learning model to focus on relevant nodes can be explained from the perspective of structural dynamics with its inherent instability. This helps to compensate the lack of explainability in attention mechanism.

The contributions of this study bring several interesting insights: First, this study provided evidence for relationship between attention mechanism, dynamics, and unstable nodes. It was found that the most relevant parts of the input data in graph neural networks are those that have ability to change the network dynamics. This study tried to explain the attention mechanism through the lens of instability analysis. Second, it was found that the collective behaviour of the imbalanced motifs in network is also determinative in changing network dynamics, and this gave evidence where we need to pay more attention to. Third, it was observed polarity-driven instabilities in hidden fractal patterns in network and this shift analytic strategy to pay more attention on hidden structures of polarity transition.

We showed that the stability analysis offers a promising solution for performing attention mechanism in a graph convolutional network faster and more efficient, by reducing computational complexity, increasing interpretability, and eliminating sensitivity to hyperparameters. Ranking stability properties of nodes makes attention models more transparent and explainable, and can be applied to a wide range of tasks including weight pruning [36], sparsification and reducing the number of non-zero weights in the network [37], making structural bias [38], etc.

The intent of these contributions is to open doors for finding explainable tools that are able to speed up the process of training in graph machine learning. We want to know if we can make the process of graph machine learning more adaptive by incorporating knowledge from stability analysis. Can prior knowledge be incorporated into graph attention network through stability analysis? How can this help to improve the accuracy of graph attention networks? If we already know from stability analysis which nodes needs attention before conducting any learning process, how can this speed up the process of aggregating information in node embedding? Can attention mechanism be replaced with stability analysis? Can we get rid of hyperparameter tuning in mechanism like biased random walk by determining transition probability based on spreading ability and stability analysis? These are the kinds of questions that we will be answering in our upcoming works.

**References**

[1] Gu, S., Pasqualetti, F., Cieslak, M. et al. Controllability of structural brain networks. Nat Commun 6, 8414 (2015). https://doi.org/10.1038/ncomms9414.

[2] Chen, C., Zhao, X., Wang, J. et al. Dynamic graph convolutional network for assembly behavior recognition based on attention mechanism and multi-scale feature fusion. Sci Rep 12, 7394 (2022). https://doi.org/10.1038/s41598-022-11206-8

[3] Zhou, P., Cao, Y., Li, M. et al. HCCANet: histopathological image grading of colorectal cancer using CNN based on multichannel fusion attention mechanism. Sci Rep 12, 15103 (2022). https://doi.org/10.1038/s41598-022-18879-1.

[4] Knyazev, B., Taylor, G. W., and Amer, M. (2019). Understanding attention and generalization in graph neural networks. In Advances in Neural Information Processing Systems (NeurIPS), arXiv:1905.02850.

[5] M. Pirani, T. Costa and S. Sundaram, "Stability of dynamical systems on a graph," 53rd IEEE Conference on Decision and Control, 2014, pp. 613-618, doi: 10.1109/CDC.2014.7039449.

[6] Meeks, Leah ; Rosenberg, David E. 2017 "High Influence: Identifying and Ranking Stability, Topological Significance, and Redundancies in Water Resource Networks" Journal of water resources planning and management, 143(6): 04017012.

[7] F. Gama, J. Bruna and A. Ribeiro, "Stability Properties of Graph Neural Networks," in IEEE Transactions on Signal Processing, vol. 68, pp. 5680-5695, 2020, doi: 10.1109/TSP.2020.3026980.

[8] F. YANG et al, "Contrastive Embedding Distribution Refinement and Entropy-Aware Attention for 3D Point Cloud Classification", 2022, arXiv:2201.11388.

[9] Li A, Huynh C, Fitzgerald Z, Cajigas I, Brusko D, Jagid J, Claudio AO, Kanner AM, Hopp J, Chen S, Haagensen J, Johnson E, Anderson W, Crone N, Inati S, Zaghloul KA, Bulacio J, Gonzalez-Martinez J, Sarma SV. Neural fragility as an EEG marker of the seizure onset zone. Nat Neurosci. 2021 Oct;24(10):1465-1474. doi: 10.1038/s41593-021-00901-w. Epub 2021 Aug 5. Erratum in: Nat Neurosci. 2022 Apr;25(4):530. PMID: 34354282; PMCID: PMC8547387.

[10] Zhang, Y., Zhang, Z., Wei, D. & Deng, Y. Centrality Measure in Weighted Networks Based on an Amoeboid Algorithm. Journal of Information and Computational Science 9, 369–376 (2012).

[11] Piraveenan, M., Prokopenko, M. & Hossain, L. Percolation centrality: quantifying graph-theoretic impact of nodes during percolation in networks. PloS one 8, e53095, https://doi.org/10.1371/journal.pone.0053095 (2013).

[12] Avena-Koenigsberger, A. et al. Path ensembles and a tradeoff between communication efficiency and resilience in the human connectome. Brain Struct Funct 222, 603–618, https://doi.org/10.1007/s00429-016-1238-5 (2017).

[13] Kwon, H., Choi, YH. & Lee, JM. A Physarum Centrality Measure of the Human Brain Network. Sci Rep 9, 5907 (2019). https://doi.org/10.1038/s41598-019-42322-7.

[14] Kitsak, M., Gallos, L., Havlin, S. et al. Identification of influential spreaders in complex networks. Nature Phys 6, 888–893 (2010). https://doi.org/10.1038/nphys1746.

[15] Sun, Y., Ma, L., Zeng, A. et al. Spreading to localized targets in complex networks. Sci Rep 6, 38865 (2016). https://doi.org/10.1038/srep38865.

[16] Zhang, C., Zhou, S., Miller, J. et al. Optimizing Hybrid Spreading in Metapopulations. Sci Rep 5, 9924 (2015). https://doi.org/10.1038/srep09924.

[17] Davis, J.T., Chinazzi, M., Perra, N. et al. Cryptic transmission of SARS-CoV-2 and the first COVID-19 wave. Nature 600, 127–132 (2021). https://doi.org/10.1038/s41586-021-04130-w.

[18] Le Treut, G., Huber, G., Kamb, M. et al. A high-resolution flux-matrix model describes the spread of diseases in a spatial network and the effect of mitigation strategies. Sci Rep 12, 15946 (2022). https://doi.org/10.1038/s41598-022-19931-w.

[19] Wang, W., Tang, M., Yang, H. et al. Asymmetrically interacting spreading dynamics on complex layered networks. Sci Rep 4, 5097 (2014). https://doi.org/10.1038/srep05097.

[20] Salnikov, V., Schaub, M. & Lambiotte, R. Using higher-order Markov models to reveal flow-based communities in networks. Sci Rep 6, 23194 (2016). https://doi.org/10.1038/srep23194.

[21] Pascual, Mercedes. "Diffusion-induced chaos in a spatial predator–prey system." Proceedings of the Royal Society of London. Series B: Biological Sciences 251 (1993): 1 - 7.

[22] Sergei Petrovskii, Bai-Lian Li, Horst Malchow, Quantification of the Spatial Aspect of Chaotic Dynamics in Biological and Chemical Systems, Bulletin of Mathematical Biology, Volume 65, Issue 3, 2003, Pages 425-446, ISSN 0092-8240, https://doi.org/10.1016/S0092-8240(03)00004-1.

[23] Nicolaou, Z.G., Case, D.J., Wee, E.B.v.d. et al. Heterogeneity-stabilized homogeneous states in driven media. Nat Commun 12, 4486 (2021). https://doi.org/10.1038/s41467-021-24459-0.

[24] Mounir Hammouche, Philippe Lutz, Micky Rakotondrabe. Robust and Optimal Output-Feedback Control for Interval State-Space Model: Application to a Two-Degrees-of-Freedom Piezoelectric Tube Actuator. Journal of Dynamic Systems, Measurement, and Control, American Society of Mechanical Engineers, 2018, 141 (2), pp.021008. ffhal-02867840.

[25] Peter Bloem, "Transformers from scratch", VU University, Aug 2019.

[26] Velickovic, P., Cucurull, G., Casanova, A., Romero, A., Lio', P., & Bengio, Y. (2018). Graph Attention Networks. ArXiv, abs/1710.10903.

[27] Bor-Sen Chen and Jung-Yuan Kung, "Robust stability of a structured perturbation system in state space models," Proceedings of the 27th IEEE Conference on Decision and Control, 1988, pp. 121-122 vol.1, doi: 10.1109/CDC.1988.194281.

[28] Saberi, M., Khosrowabadi, R., Khatibi, A. et al. Topological impact of negative links on the stability of resting-state brain network. Sci Rep 11, 2176 (2021). https://doi.org/10.1038/s41598-021-81767-7.

[29] M. Golubitsky, et all, Symmetry and Bifurcation in Biology, BIRS, June, 2003.

[30] Ruzzenenti, F.; Garlaschelli, D.; Basosi, R. Complex Networks and Symmetry II: Reciprocity and Evolution of World Trade. Symmetry 2010, 2, 1710-1744. https://doi.org/10.3390/sym2031710.

[31] Goirand, F., Le Borgne, T. & Lorthois, S. Network-driven anomalous transport is a fundamental component of brain microvascular dysfunction. Nat Commun 12, 7295 (2021). https://doi.org/10.1038/s41467-021-27534-8.

[32] M. A. Broussard, Diagram of lamellate antenna, 27 March 2016, based on File:Ten-lined June beetle Close-up.jpg.

[33] Lazaro M. Sanchez-Rodriguez, Yasser Iturria-Medina, Pauline Mouches, Roberto C. Sotero, Detecting brain network communities: Considering the role of information flow and its different temporal scales, NeuroImage, Volume 225, 2021, 117431, ISSN 1053-8119, https://doi.org/10.1016/j.neuroimage.2020.117431.

[34] Fallani Fde V, Costa Lda F, Rodriguez FA, Astolfi L, Vecchiato G, Toppi J, Borghini G, Cincotti F, Mattia D, Salinari S, Isabella R, Babiloni F. A graph-theoretical approach in brain functional networks. Possible implications in EEG studies. Nonlinear Biomed Phys. 2010 Jun

3;4 Suppl 1(Suppl 1):S8. doi: 10.1186/1753-4631-4-S1-S8. PMID: 20522269; PMCID: PMC2880805.

[35] Rosvall, M., Esquivel, A., Lancichinetti, A. et al. Memory in network flows and its effects on spreading dynamics and community detection. Nat Commun 5, 4630 (2014). https://doi.org/10.1038/ncomms5630.

[36] Wang, Jialin, Rui Gao, H. Zheng, Hao Zhu and Chenyun Shi. "SSGCNet: A Sparse Spectra Graph Convolutional Network for Epileptic EEG Signal Classification." (2022).

[37] Palcu L-D., Supuran M., Lemnaru C., Dinsoreanu M., Potolea R., Muresan R.C. (2019), Breaking the interpretability barrier - a method for interpreting deep graph convolutional models. International Workshop NFMCP in conjunction with ECML-PKDD 2019, Wurzburg, Germany.

[38] Patil, Akshay Gadi, Manyi Li, Matthew Fisher, Manolis Savva and Hao Zhang. "LayoutGMN: Neural Graph Matching for Structural Layout Similarity." 2021 IEEE/CVF Conference on Computer Vision and Pattern Recognition (CVPR) (2020): 11043-11052.

## Appendix

In symmetry-breaking stability analysis, following calculations is the process of computing spreading ability of each node based on Equation (7). Figure 17 shows one example of path's weights traversed during a single random walk

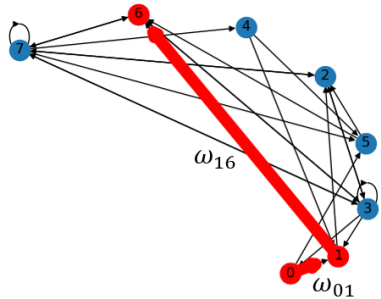

Fig. 17. Path's weights traversed from node 0 during a single random walk

$NSTC_0 =$

$$\frac{\left((\omega_{01} \times \omega_{12}) + (\omega_{01} \times \omega_{16}) + (\omega_{05} \times \omega_{52}) + (\omega_{05} \times \omega_{56})\right)}{N_0} = \frac{\left((1 \times 1) + (1 \times 1) + (1 \times 1) + (1 \times 1)\right)}{4} = 1$$

$NSTC_1 =$

$$\frac{((\omega_{12} \times \omega_{23}) + (\omega_{12} \times \omega_{27}) + (\omega_{16} \times \omega_{63}) + (\omega_{16} \times \omega_{67}))}{N_1} = \frac{((1 \times 1) + (1 \times 1) + (1 \times 1) + (1 \times 1))}{4} = 1$$

$NSTC_2$
$$= \frac{((\omega_{23} \times \omega_{30}) + (\omega_{23} \times \omega_{31}) + (\omega_{23} \times \omega_{36}) + (\omega_{23} \times \omega_{37}) + (\omega_{27} \times \omega_{73}) + (\omega_{27} \times \omega_{74}) + (\omega_{27} \times \omega_{75}) + (\omega_{27} \times \omega_{76}))}{N_2}$$
$$= \frac{\begin{pmatrix} (1 \times -2.748) + (1 \times 1.3083) + (1 \times -2.6331) + (1 \times -0.9492) + \\ (1 \times -191.8224) + (1 \times -5.2346) + (1 \times -3.6549) + (1 \times -1.7819) \end{pmatrix}}{8} = -25.9395$$

$NSTC_3 =$

$$\frac{\begin{pmatrix}(\omega_{30} \times \omega_{01}) + (\omega_{30} \times \omega_{05}) + (\omega_{31} \times \omega_{12}) + (\omega_{31} \times \omega_{16}) + (\omega_{32} \times \omega_{27}) + (\omega_{36} \times \omega_{67}) + \\ (\omega_{37} \times \omega_{72}) + (\omega_{37} \times \omega_{74}) + (\omega_{37} \times \omega_{75}) + (\omega_{37} \times \omega_{76})\end{pmatrix}}{N_3} =$$

$$\frac{\begin{pmatrix}(-2.748 \times 1) + (-2.748 \times 1) + (1.3083 \times 1) + (1.3083 \times 1) + (-4.2614 \times 1) + (-2.6331 \times 1) + \\ (-0.9492 \times -49.4899) + (-0.9492 \times -5.2346) + (-0.9492 \times -3.6549) + (-0.9492 \times -1.7819)\end{pmatrix}}{10}$$
$$= 4.7331$$

$NSTC_4 =$

$$\frac{((\omega_{41} \times \omega_{12}) + (\omega_{41} \times \omega_{16}) + (\omega_{45} \times \omega_{52}) + (\omega_{45} \times \omega_{56}))}{N_4} = \frac{((1 \times 1) + (1 \times 1) + (1 \times 1) + (1 \times 1))}{4} = 1$$

$NSTC_5 =$

$$\frac{((\omega_{52} \times \omega_{23}) + (\omega_{52} \times \omega_{27}) + (\omega_{56} \times \omega_{63}) + (\omega_{56} \times \omega_{67}))}{N_5} = \frac{((1 \times 1) + (1 \times 1) + (1 \times 1) + (1 \times 1))}{4} = 1$$

$NSTC_6 =$

$$\frac{((\omega_{63} \times \omega_{30}) + (\omega_{63} \times \omega_{31}) + (\omega_{63} \times \omega_{32}) + (\omega_{63} \times \omega_{37}) + (\omega_{67} \times \omega_{72}) + (\omega_{67} \times \omega_{73}) + (\omega_{67} \times \omega_{74}) + (\omega_{67} \times \omega_{75}))}{N_2}$$
$$= \frac{\begin{pmatrix}(1 \times -2.748) + (1 \times 1.3083) + (1 \times -4.2614) + (1 \times -0.9492) + \\ (1 \times -49.4899) + (1 \times -191.8224) + (1 \times -5.2346) + (1 \times -3.6549)\end{pmatrix}}{8} = -32.1065$$

$NSTC_7 =$

$$\frac{\begin{pmatrix}(\omega_{72} \times \omega_{23}) + (\omega_{73} \times \omega_{30}) + (\omega_{73} \times \omega_{31}) + (\omega_{73} \times \omega_{32}) + (\omega_{73} \times \omega_{36}) + (\omega_{74} \times \omega_{41}) + \\ (\omega_{74} \times \omega_{45}) + (\omega_{75} \times \omega_{52}) + (\omega_{75} \times \omega_{56}) + (\omega_{76} \times \omega_{63})\end{pmatrix}}{N_7} =$$

$$\frac{\begin{pmatrix}(-49.4899 \times 1) + (-191.8224 \times -2.748) + (-191.8224 \times 1.3083) + (-191.8224 \times -4.2614) + \\ (-191.8224 \times -2.6331) + (-5.2346 \times 1) + (-5.2346 \times 1) + (-3.6549 \times 1) + (-3.6549 \times 1) + (-1.7819 \times 1)\end{pmatrix}}{10}$$
$= 152.9635$